# APP: A* Post-Processing Algorithm for Robots with Bidirectional Shortcut and Path Perturbation

Yong Li[1,2,*], *Member*, *IEEE*, and Hui Cheng[2], *Member*, *IEEE*

*Abstract*— Paths generated by A* and other graph-search-based planners are widely used in the robotic field. Due to the restricted node-expansion directions, the resulting paths are usually not the shortest. Besides, unnecessary heading changes, or zig-zag patterns, exist even when no obstacle is nearby, which is inconsistent with the human intuition that the path segments should be straight in wide-open space due to the absence of obstacles. This article puts forward a general and systematic post-processing algorithm for A* and other graph-search-based planners. The A* post-processing algorithm, called APP, is developed based on the costmap, which is widely used in commercial service robots. First, a bidirectional vertices reduction algorithm is proposed to tackle the asymmetry of the path and the environments. During the forward and backward vertices reduction, a thorough shortcut strategy is put forward to improve the path-shortening performance and avoid unnecessary heading changes. Second, an iterative path perturbation algorithm is adopted to locally reduce the number of unnecessary heading changes and improve the path smoothness. Comparative experiments are then carried out to validate the superiority of the proposed method. Quantitative performance indexes show that APP outperforms the existing methods in planning time, path length as well as the number of unnecessary heading changes. Finally, field navigation experiments are carried out to verify the practicability of APP.

## I. INTRODUCTION

### A. Motivation

As a typical service robot, the cleaning robot is used to clean the solid and liquid wastes on the ground [1]. Household cleaning robots have been widely used, while the commercial ones working in offices, lobbies, airports, hospitals and schools are still at an early stage [2][3]. There are still many problems to be solved, and one of them is finding a practical global path.

Common path planning algorithms can be classified into four types [4]: graph-search-based planners, sampling-based methods, interpolating-curve-based and optimization-based ones. Among them, the graph-search-based methods represented by A* are widely used in the robotic field [5][6].

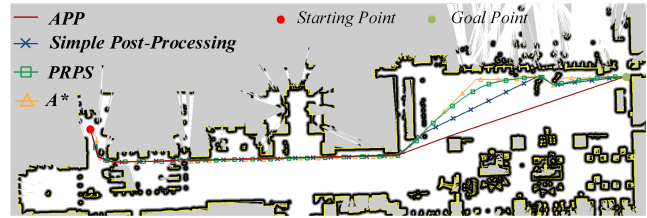

Fig. 1. Comparative experimental results of **A**\* **P**ost-**P**rocessing (APP) algorithm, simple post-processing algorithm [12-14] and **P**ath **R**elaxation and **P**ath **S**moothing (PRPS) algorithm [15], with the experimental environment the lobby of a hotel. Compared with the original path of A*, traditional post-processing methods use a unidirectional and simple iterative shortcut technique and only constrained path-shortening performance can be achieved. PRPS aims to shorten path length as well as improve path smoothness. However, path-shortening performance is not satisfactory even when the computation time for the post-processing process is three times that for the path-finding process. In APP, a bidirectional and thorough-shortcut-based vertices reduction algorithm is put forward to guarantee improved path-shortening performance.

The A* algorithm expands nodes in 4-neighbor or 8-neighbor for 2D applications [7], which means that the node-expansion process is constrained to be horizontal, vertical, or 45° tilted. Due to the limited node-expansion directions, the resulting path is very likely not to be the shortest. Besides, the resulting path usually has some unnecessary heading changes, which affects the smoothness and is not comply with the common sense of human beings: the path should be straight in wide-open space due to the absence of obstacles. Eliminating unnecessary heading changes and making the path understandable are important, however, to which little attention has been paid in the literature.

Shortening path length, eliminating unnecessary heading changes and improving path smoothness are essential for commercializing robots in unstructured and semi-structured environments, and are the main focus of this article.

### B. Related Work

Approaches to the above problems of traditional A* can be classified into two categories: A*-variant and post-processing-based methods.

Theta* is a typical variant of A* that is developed to shorten the path length [8]. During the expansion of current node $s$, its visible neighbor $s^{'}$ will share the same father node $s.father$ with $s$ if there is no obstacle between $s.father$ and $s^{'}$. To improve the computational efficiency of Theta* in 3D applications, Nash et al. introduce Lazy Theta*, which performs a line-of-sight check only per node in the closed set [9]. Theta* and Lazy Theta* are developed based on the grid map with only two states: free and obstacle. Choi et al. extend Theta* to accommodate the costmap since it can represent more information than the simple binary grid map[10]. Besi-des, Kim et al. apply a limit-cycle circle set to Theta* with considering the vehicle's heading angle [11]. The approaches mentioned above can reduce some unnecessary nodes during the node-expansion process, which is helpful for path shortening. However, they are not fundamental solutions as they do not have an effective mechanism for unnecessary-heading-changes-eliminating, and the zig-zag symptom still exists, which can be seen in Fig.10, 10 and 7 of [8], [10] and [11], respectively.

[1]Guangzhou Shiyuan Electronic Technology Co. Ltd., Guangzhou 510300, China.
[2]School of Computer Science and Engineering, Sun Yat-sen University, Guangzhou 510006, China.
*Corresponding author: Yong Li (e-mail: liyong2018@zju.edu.cn).

In the literature, most of the post-processing methods for graph-search-based dense paths are simple and straightforward [12-14]. The algorithm starts at the first vertex $p_0$ of the path $\mathcal{P}$ and advances along $\mathcal{P}$ to find the first vertex $p_k$ that is not visible with $p_0$. As soon as $p_k$ is found, the post-processing for $p_0$ stops, and the vertices between $p_0$ and $p_{k-1}$ are removed. Then the iterations for $p_{k-1}$ continue till the end of $\mathcal{P}$. As seen from Fig. 1 of this article and Fig. 4 in [14], such a simple shortcut strategy can only achieve constrained path-shortening performance. A. Richardson and E. Olson put forward an iterative path post-processing method for robotic planning [15], which is considered state-of-the-art for dense paths generated by graph-search-based planners. The post-processing process has two main steps: Firstly, the path is relaxed by moving path points individually down the cost surface to increase clear-up. Secondly, the path will be smoothed and shortened around the cost surface. However, the path-smoothing process is very time-consuming: as seen in Table I of [15], the computation time for post-processing is roughly three times that for the path-finding, which restricts its application on low-cost computation units.

Post-processing approaches for sparse paths generated by sampling-based planners, such as RRT, PRM, and FMT have also been put forward [16-19]. Geraerts et al. propose a shortcut technique that removes redundant vertices between two randomly chosen vertices within a predefined threshold window [16]. Similar practices can be seen in post-processing algorithms (Functions "reduceVertices" and "shortcutPath") in OMPL [17]. The shortcut performances of such practices are restricted by the window size and the generation of random numbers. Heiden et al. propose a hybrid strategy combining gradient-based optimization and shortcut [18], which is considered state-of-the-art for sparse paths generated by sampling-based planners. In the shortcut process, the current vertex is considered a "necessary vertex" if its predecessor is not visible with its successor, and an acyclic graph based on all necessary vertices will then be built to find the shortest path. However, this approach can not be directly applied to dense paths generated by A* or other graph-search-based planners: the predecessor of a vertex for a dense path is visible with its successor most of the time, and it is difficult to find the necessary vertices to build a valid graph.

A practical post-processing algorithm for A* and other graph-search-based planners should be able to shorten path length, eliminate unnecessary heading changes and improve path smoothness, with little extra computation burden. As far as the authors can see, there are no systematic approaches in the literature.

*C. Contributions*

This article proposes a fast and practical post-processing algorithm for dense paths generated by A* and other graph-search-based planners. The main contributions are:

(1) A general and systematic A* post-processing algorithm is proposed for robotic applications, which can effectively shorten path length, avoid unnecessary heading changes, and improve the smoothness of the path;

(2) A bidirectional vertices reduction approach based on the costmap and thorough shortcut is proposed, which guarantees balanced path length and path safety in grid space; Besides, a path perturbation algorithm is introduced to improve the smoothness of the path in the sub-grid further;

(3) Comparative experiments with several typical maps of a commercial cleaning robot are carried out, and quantitative performance indexes are introduced to verify the superiority of the proposed method.

(4) The proposed method requires a little extra computation burden and is practical even on low-cost computation units, which means it can be applied to offline global path planning as well as online local path replanning. Field navigation experiments are conducted to verify its practicability.

The rest of the article is organized as follows: Sec. II is about the problem description, followed by Sec. III presenting the methodology. Then comparative and field experimental results are described in Sec. IV. Conclusions are drawn in Sec. V.

## II. PROBLEM DESCRIPTION

For the given starting point $p_s$ and goal point $p_g$, the classic A* algorithm can generate a dense path $P_0$, which consists of consecutive vertices $P_0 = \{p_0 = p_s, p_1, p_2, \ldots, p_{n-2}, p_{n-1} = p_g\}$. The Euclidean distances between two consecutive vertices are the map resolution $\xi$ (or the square root of $\xi$). The map $\mathcal{M}$ used in the path planning and post-processing is a costmap with inflated obstacles [20-22]. The costs of obstacles and unknown grids are 254 and 255, respectively. For any free grid cell $c_i \in \mathcal{M}$ with its nearest obstacle grid $c_j$, its cost $\mathcal{C}_i$ is calculated as [20]:

Unknown　　　Obstacle　　　Inflated grids

Fig. 2. Part of a costmap for a commercial cleaning robot.

$$\mathcal{C}_i = \begin{cases} 253, & if \ 0 < d_{ij} \leq r_1 \\ 253 e^{-w(d_{ij}-r_1)}, & if \ r_1 < d_{ij} \leq r_2 \\ 0, & if \ r_2 < d_{ij} \end{cases} \quad (1)$$

where $d_{ij}$ is the Euclidean distance between $c_i$ and $c_j$, $r_1$ the inscribed radius of the robot, and $r_2$ the cut-off distance for grid inflation.

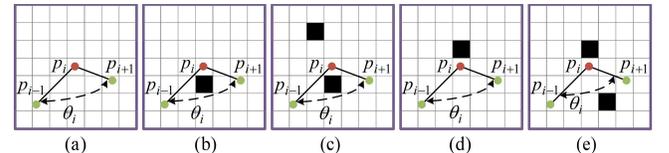

(a)　　(b)　　(c)　　(d)　　(e)

Fig. 3. Necessary (*b, c, e*) and unnecessary (*a, d*) heading changes at vertex $p_i$. The angle between two consecutive path segments $\overrightarrow{p_i p_{i-1}}$ and $\overrightarrow{p_i p_{i+1}}$ is $\theta_i$. The black grids represent obstacles around $p_i$.

Daniel et al. define "unnecessary heading changes" as heading changes that occur in free space rather than the corners of obstacle grids [8]. This definition is not rigorous: as seen from Fig.3(d), heading change at vertex $p_i$ occurs around the obstacle but is obviously unnecessary. In this article, the set of vertices with unnecessary heading change $\mathcal{H}_\mathcal{P}$ for path $\mathcal{P}$ is defined as:

$$\mathcal{H}_\mathcal{P} = \mathcal{H}1_\mathcal{P} + \mathcal{H}2_\mathcal{P}$$
$$\mathcal{H}1_\mathcal{P} = \{p_i \in \mathcal{P}: \theta_{p_i} < \pi \ \& \ \{\mathcal{O} = \emptyset, \mathcal{O} \in \mathcal{W}_i\}\} \quad (2)$$
$$\mathcal{H}2_\mathcal{P} = \{p_i \in \mathcal{P}: \theta_{p_i} < \pi \ \& \ \{\forall \mathcal{O}_i \in \mathcal{W}_i, \mathcal{O}_i \notin \theta_{p_i}\}\}$$

where $p_i$ is any vertex of $\mathcal{P}$, $\theta_{p_i}$ the angle between $\overrightarrow{p_i p_{i-1}}$ and $\overrightarrow{p_i p_{i+1}}$, $\mathcal{W}_i$ an obstacle-checking window centered at $p_i$, $\mathcal{O}$ the set of obstacles within $\mathcal{W}_i$. $\mathcal{O}_i$ is considered as "not within" $\theta_{p_i}$ if:

$$\mathcal{O}_i \notin \theta_{p_i} = \neg(\mathcal{M}_{13} \geq 0 \ \& \ \mathcal{M}_{23} \leq 0) \quad (3)$$

where $\mathcal{M}_{13} = \overrightarrow{p_i p_{i-1}} \times \overrightarrow{p_i \mathcal{O}_i}$, $\mathcal{M}_{23} = \overrightarrow{p_i p_{i+1}} \times \overrightarrow{p_i \mathcal{O}_i}$. $\mathcal{W}_i$ can be determined according to the robot's configuration space and is represented as the purple box in this paper in Fig. 3.

According to Equations (2) and (3), if there is not any obstacle around $p_i$, the heading change at $p_i$ is considered unnecessary, corresponding to Fig. 3(a) and $\mathcal{H}1_\mathcal{P}$. Different from Daniel et al.'s definition, the heading change in Fig. 3(d) is also considered unnecessary as it has nothing to do with the obstacle around $p_i$, corresponding to $\mathcal{H}2_\mathcal{P}$.

The objective of APP is to shorten path length, eliminate unnecessary heading changes and improve path smoothness, with considering computation efficiency.

---

**Algorithm 1: Procedure for APP**

**Input**: A* Path $P_0$, Cycle Num $n_{iter}, n_{sm}$, Res $\xi_{inte}$
Threhold $\varepsilon_{len}, \varepsilon_{cost}, \varepsilon_{sm}$
**Output**: SmoothedPath $P_{sm}$

1 **MainFlow**:
2    $LastPathLen \leftarrow +\infty$;
3    $P_{spa} \leftarrow \emptyset, P_{raw} \leftarrow P_0$;
4    $PathLen \leftarrow P_{raw}.len(), LoopCnt \leftarrow 0$;
5    **while** $|LastPathLen - PathLen| > \varepsilon_{len} \wedge LoopCnt <= n_{iter}$ **do**
6      $LoopCnt \leftarrow LoopCnt + 1$;
7      $LastPathLen \leftarrow PathLen$;
8      $P_{in} \leftarrow P_{raw}$;
9      $P_{for} \leftarrow \textbf{ReduceVertices}(P_{in}, \varepsilon_{cost})$;
10      $P_{spa} \leftarrow P_{for}$;
11      /* shortcut backward */
12      $P_{in} \leftarrow P_{raw}.reverse()$;
13      $P_{back} \leftarrow \textbf{ReduceVertices}(P_{in}, \varepsilon_{cost})$;
14      **if** $P_{for}.len() > P_{back}.len()$ **then**
15        $P_{spa} \leftarrow P_{back}.reverse()$;
16      **end if**
17      $P_{dens} \leftarrow \textbf{InterpolatePath}(P_{spa}, \xi_{inte})$;
18      $P_{sm} \leftarrow \textbf{PerturbPath}(P_{dens}, n_{sm}, \varepsilon_{cost}, \varepsilon_{sm})$;
19      $P_{raw} \leftarrow P_{sm}$;
20      $PathLen \leftarrow P_{sm}.len()$;
21    **end while**
22    **return** $P_{sm}$;
23 **End MainFlow**

---

## III. METHODOLOGY

### A. Algorithm Flow of APP

The procedure for APP is shown in Algo. 1. The input and output are raw path $P_{raw}$ (initialized as A* path $P_0$) and the resulting smoothed path $P_{sm}$, respectively. APP is an iterative algorithm, and when the cycle time $LoopCnt$ for the outer loop exceeds the iteration number threshold $n_{iter}$, or the variation of $P_{sm}$ in two consecutive cycles is within the path length threshold $\varepsilon_{len}$, APP stops iteration.

For each outer loop iteration, the vertices reduction process executes firstly from forward and then backward with the resulting sparse path $P_{for}$ and $P_{back}$, respectively, among which the shorter one is donated as $P_{spa}$. $P_{spa}$ is a sparse path and will then be linearly interpolated to make it a dense path ($P_{dens}$) again. $P_{dens}$ will be perturbed to improve the smoothness and eliminate unnecessary heading changes. The resulting smoothed path $P_{sm}$ will be adopted as the input path $P_{raw}$ for the following cycle. Detailed elaboration for bidirectional vertices reduction, path interpolation and path perturbation are as follows.

---

**Algorithm 2: Procedure for ReduceVertices**

1 **ReduceVertices**($P_{in}, \varepsilon_{cost}$):
2    $P_{out} \leftarrow \emptyset$;
3    $P_{out} \leftarrow P_{out} \cup P_{in}.front()$;
4    $p_0 \leftarrow P_{out}.back()$;
5    $AnchorIndex \leftarrow 0$;
6    $Candidates \leftarrow \emptyset$;
7    **while** $AnchorIndex < P_{in}.size() - 1$ **do**
8      **for** $i \leftarrow AnchorIndex + 1$ **to** $P_{in}.size() - 1$ **do**
9        **if** $\textbf{LineofSight}(p_0, P_{in}[i], \varepsilon_{cost})$ **then**
10          $Candidates \leftarrow Candidates \cup \{i, P_{in}[i]\}$;
11        **end if**
12      **end for**
13      **if** $Candidates = \emptyset$ **then**
14        $Candidates \leftarrow Candidates \cup \{AnchorIndex + 1, P_{in}[AnchorIndex + 1]\}$;
15      **end if**
16      /* shortcut to get the last visible vertex from Candidates */
17      $Last \leftarrow Candidates.back()$;
18      $P_{out} \leftarrow P_{out} \cup Last.second()$;
19      $p_0 \leftarrow Last.second()$;
20      $AnchorIndex \leftarrow Last.first()$;
21      $Candidates \leftarrow \emptyset$;
22    **end while**
23    **return** $P_{out}$;
24 **End ReduceVertices**

---

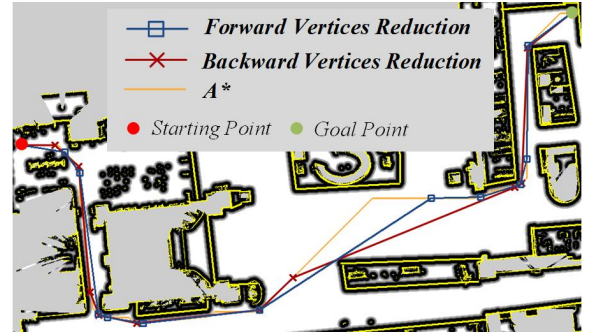

Fig. 4. Comparative experimental results between forward and backward vertices reduction for the first iteration. Due to the complexity and asymmetry of the environment, $P_{for}$ differs from $P_{back}$ in topological structure.

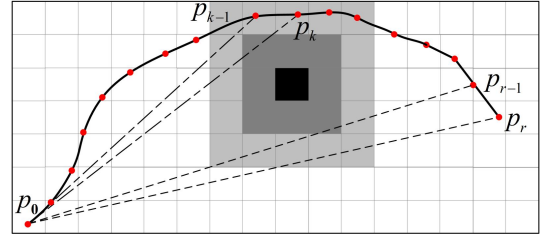

Fig. 5. Vertices reduction for path $\mathcal{P}$ (solid black line with the red dots representing vertices) based on the costmap. It can be noticed that the vertices are not necessarily located at the center of the girds, which is due to the float operation in path perturbation. The black square represents the obstacle, and the gray ones stand for the inflated grids. The cost of the eight medium-gray cells is defined as the threshold cost $\varepsilon_{cost}$ for the line-of-sight check. Starting at vertex $p_0$ and advancing along path $\mathcal{P}$, $p_k$ is the first vertex not visible with $p_0$, with $p_r$ the last one visible with $p_0$.

## B. Reduce Vertices

The procedure for reducing vertices is shown in Alg. 2. The introduction of the bidirectional shortcut process is to tackle the complexity and asymmetry of the environment. As seen from Fig. 4, the resulting path of forward shortcut $P_{for}$ differs from that of backward $P_{back}$ in topological structure as well as path length, and the shorter one will be selected. The vertices reduction process has two distinctive features, which benefit for satisfying the set-out demands of robotics.

First, a thorough shortcut algorithm is put forward to avoid local optimum. As seen from Fig. 5, for a dense path $\mathcal{P} = \{p_0, p_1, \ldots, p_{n-2}, p_{n-1}\}$, traditional shortcut strategy [12-14] selects $p_0$ as the current anchor vertex. Then it advances along $\mathcal{P}$ to find the first vertex $p_k$ that is not visible with $p_0$. Vertices between $p_0$ and $p_{k-1}$ are unnecessary and can be removed from $\mathcal{P}$. $p_{k-1}$ is then selected as the current anchor vertex and repeat this procedure until it reaches the end of $\mathcal{P}$. Such practice is suboptimal as it does not have a global view. Fortunately, as seen from Fig. 5 and Algo. 2, an improved strategy is adopted for APP. The vertex reduction does not care about the vertexes that are not visible with the current vertex $p_0$. Instead, the algorithm advances along $\mathcal{P}$ (Line 8-12 in Algo. 2) to find each vertex $p_j$ visible with $p_0$ and stores it into the set *Candidates* (Line 10 in Algo. 2). The last vertex in *Candidates* is the farthest one visible with $p_0$ (vertex $p_r$ in Fig. 5) and vertices between it and $p_0$ are unnecessary and can be removed (Line 17-18 in Algo. 2). $p_r$ will be selected as the new current anchor vertex to iterate through $\mathcal{P}$. The resulting vertex set, $P_{out}$, is a minimum composition of $\mathcal{P}$.

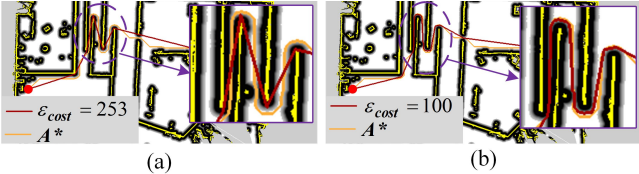

Fig. 6. Comparative path planning results based on (a) the traditional line-of-sight check [8-10] and (b) the one proposed in this article.

Second, a costmap-based line-of-sight check is introduced to improve the safety of the path. In video games or simple robotic applications, each grid of the map has only two states: free and obstacle. The criterion for the line-of-sight-check-based shortcut is that the line should not traverse any obstacle grids [8-10], resulting in that some vertices of the resulting path are unavoidably adjacent to obstacles and sharp angles occurs, as seen from Fig. 6(a). It is normal for a virtual hero in video games that some vertices of the path are adjacent to obstacle girds but is unacceptable for robots as it greatly reduces navigation safety considering localization and path following errors. To solve this problem, we extend the traditional line-of-sight check to accommodate the costmap that the costs of the grids traversed by the line should not exceed a predefined cost threshold $\varepsilon_{cost}$, a much smaller value than the obstacle cost 254, as can be seen in Fig. 6(b). Besides, when the original A* path traverses a narrow passage, the costs of some vertices can be higher than $\varepsilon_{cost}$. For any vertex with a cost higher than $\varepsilon_{cost}$, the line-of-sight check between it and any other vertices will always fail. To solve the problem, when a vertex can not find any vertex that satisfies the line-of-sight condition, its successor will be added as a visible candidate, as seen in Lines 13-15 in Algo. 2. This means that, for APP, the path will only be shortened and tightened in low-cost grids, and the ones with a higher cost than $\varepsilon_{cost}$ remain unchanged, which makes a guaranteed balance between path length and path safety.

## C. Linear Interpolation

As seen from Fig. 4, after the vertices reduction for the first iteration, the original dense A* path is represented as a sparse path $P_{spa}$. Due to the effective vertices reduction algorithm proposed in this article, $P_{spa}$ is already a minimum set, and its vertices can not be reduced further. Linear interpolation aims to make $P_{spa}$ a dense path again so that the iterative post-processing process can continue. To avoid a heavy computation burden, the interpolation resolution $\xi_{inte}$ can be set larger than the map resolution $\xi$.

---
**Algorithm 3: Procedure for PerturbPath**

1  PerturbPath($P_{dens}, n_{sm}, \varepsilon_{cost}, \varepsilon_{sm}$):
2    $P_{sm} \leftarrow \emptyset$;
3    for $i \leftarrow 0$ to $n_{sm}$ do
4      $bCut \leftarrow false$;
5      for $j \leftarrow 1$ to $P_{dens}.size() - 2$ do
6        if $LineofSight(p_{j-1}, p_{j+1}, \varepsilon_{cost})$ then
7          $p_{tmp} \leftarrow (p_{j-1} + p_{j+1})/2$;
8          if $Cost[p_{tmp}] < \varepsilon_{cost} \wedge |p_{tmp} - p_j| > \varepsilon_{sm}$ then
9            $p_j \leftarrow p_{tmp}$;
10           $bCut \leftarrow true$;
11         end if
12       end if
13     end for
14     if $bCut = false$ then
15       break;
16     end if
17   end for
18   $P_{sm} \leftarrow P_{dens}$;
19   return $P_{sm}$;
20 End PerturbPath

---

## D. Perturb Path

Path perturbation runs in conjunction with the linear interpolation to improve the smoothness of the path and eliminate unnecessary heading changes in the sub-grid. Specific vertices are perturbed, and the total number of vertices will not be changed.

As seen from Algo. 3, for any vertex $p_i$, the cost for the middle point $p_{tmp}$ of its predecessor $p_{i-1}$ and successor $p_{i+1}$ is $Cost_{p_{tmp}}$. If $p_{i-1}$ is visible with $p_{i+1}$, $Cost_{p_{tmp}}$ smaller than cost threshold $\varepsilon_{cost}$, and the distance between $p_i$ and $p_{tmp}$ is larger than distance threshold $\varepsilon_{sm}$, $p_i$ will be replaced by $p_{tmp}$ (Line 6-12 in Algo. 3). Besides, a branch-and-bound method is adopted to improve the computation efficiency: If no vertex is perturbed for the current cycle, i.e., the state variable $bCut$ remains unchanged, the path perturbation process ends (Line 14-16 in Algo. 3).

The path perturbation algorithm is simple but effective in improving the smoothness of the path in the sub-grid. Furthermore, unnecessary heading changes can also be eliminated through the iterative path perturbation process.

## IV. EXPERIMENTS

### A. Comparative experiments

To verify the effectiveness of the proposed methods, comparative experiments are carried out. The computation unit used in the experiments is an industrial computer with the CPU i7-10700@2.9Hz×16 and RAM of 32 GB. The maps u-

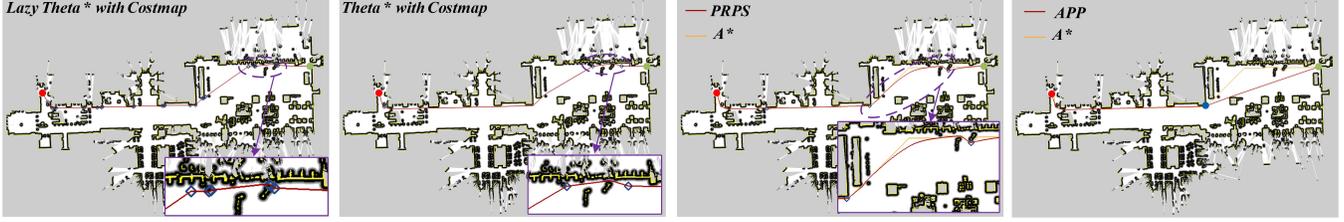

Fig. 7. Comparative experimental results in experiment I. Parts of the paths are enlarged to demonstrate the details better.

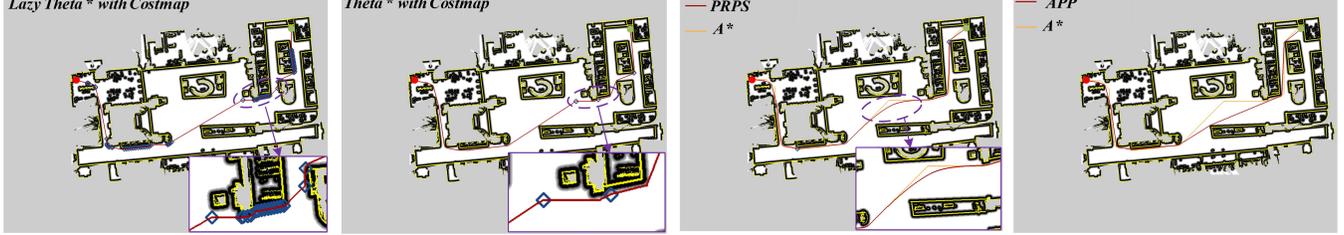

Fig. 8. Comparative experimental results in experiment II. Parts of the paths are enlarged to demonstrate the details better.

sed in the experiments arise from the actual working environments of the commercial cleaning robots "CLEAN III," produced by CVTE and the map resolutions are all 0.05m. The following four algorithms are compared:

**Theta\* with Costmap** [10]: Daniel et al. propose the Basic Theta\* [8] and Choi et al. extend it to accommodate the costmap [10], which is considered state-of-the-art for A\* variants in terms of path-shortening. During the node-expansion process, the arithmetic-mean-based method is adopted to calculate the edge cost as it has better computational efficiency than the weighted-mean-based one [10].

**Lazy Theta\*** [9] **with Costmap**: Lazy Theta\* is a typical variant of Theta\*. According to the author's experimental results [9], for 3D simulation environments with obstacles less than 30%, Lazy Theta A\* has better computational efficiency and comparative performance with Theta A\* in path length. Similar to Choi's work [10], we extend Lazy Theta\* to accommodate the costmap.

**PRPS:** Path post-processing method based on path relaxation and path smoothing proposed by A. Richardson et al. [15]. Path relaxation aims to move path points down the cost surface to increase clear-up, while path smoothing brings point $i$ closer to the line segment defined by $i-1$ and $i+1$ to shorten the path. In PRPS, the path planning process is based on Dijkstra and a simple binary obstacle map. To make the comparison fair, PRPS shares the same path planner, A\* with costmap, with the other three methods during the comparative experiments. Richardson et al. do not provide detailed instructions on parameter-turning in their work, especially the upper bound of iteration for path relaxation. As seen from the experimental results in Table I [15], the computation time for post-processing (relaxation and smoothing) is roughly three times that for path planning, with the map resolution 0.05m. In this article, when the upper bound of iteration for path relaxation is selected as 3000 in Experiment I, the ratio of computation time between post-processing and path planning is roughly the same as that in [15]. And the same upper bound is adopted in Experiments I-IV.

**APP**: The method proposed in this article. The outer loop iteration number $n_{iter}$ and path perturbation iteration number $n_{sm}$ are set to be 5 and 20, respectively. The linear interpolation resolution $\xi_{inte}$ is set to be 20, and the path length threshold value for outer loop $\varepsilon_{len}$ can be set as a value proportional to the path length. In this article, it is set to be 1. Path perturbation distance threshold $\varepsilon_{sm}$ is set to be 0.01 to achieve a better smoothing performance in the sub-grid.

Choi et al. introduce the basic idea for their work [11], but some detailed information about the experiments is not included. In this article, the costmap is obtained with Equ. 1, where $\varepsilon_{cost}$, $r_1$, $r_2$ and $w$ are set to be 100, 0.23, 0.5 and 6, respectively. Besides, for A\*, Theta\* and Lazy Theta\*, the heuristic cost $h(s, p_g)$ and edge cost $c(s, s')$ are set to be proportional to Manhattan and Euclidean distances, respectively, with the coefficient 50. To avoid drawbacks of the traditional line-of-sight check, as can be seen in Fig. 6(a), the line-of-sight check for Theta\* and Lazy Theta\* are upgraded to the methods in APP that the threshold-based criterion is adopted, with the cost threshold $\varepsilon_{cost}$ 100.

To quantify the performances of different methods, the following performance indexes are introduced:

- $t = (1/N_t) \sum_{i=0}^{N_t-1} t_i$, the average planning time (for APP and PRPS, $t$ is the total planning time including A\* and post-processing) for $N_t$ consecutive planning cycles, is used to evaluate the average computational efficiency. $N_t$ is set to be 100 in the experiments;
- $l$ the path length in grids, is used to evaluate the path length cost;
- $\mathcal{C} = (1/n) \sum_{i=0}^{n-1} \mathcal{C}_i$ with $\mathcal{C}_i$ the cost of vertex $p_i$, is used to evaluate the safety of the path;
- $\mathcal{N}_\theta$ the number of unnecessary heading changes $\mathcal{H}_\mathcal{P}$, is used to evaluate the straightness and rationality of the path.
- $S = (1/n) \sum_{i=0}^{n-1} |\pi - \theta_i|$ with $\theta_i$ the angle between two consecutive path segments, is used to evaluate the smoothness of the path;

Besides, the density of the path vertices has a significant impact on $\mathcal{C}$ and $S$. To make the comparison fair and correct, the resulting paths of the four methods are linearly interpolated with the map resolution 0.05 before calculating $\mathcal{C}$ and $S$.

**Experiment I**: The experimental environment is the lobby of a hotel with the map grid size of 1800 × 1200. Comparative experimental results and performance indexes are shown in Fig. 7 and Table I, respectively. The planning efficiency of APP is slightly better than Lazy Theta* and Theta* in this simple environment. However, due to the adoption of the superior vertices reduction algorithm, the latter part of the path for APP is reduced to a straight line (the path segment between the blue vertex and goal point in Fig. 7). And significant performance improvements in path length, unnecessary heading change, path cost as well as path smoothness have been achieved. Actually, the basic logic for vertices reduction in Lazy Theta* and Theta* is similar to that in simple post-processing, which can only achieve constrained shortcut performance, as can be seen from Fig. 1 and 7. Besides, for PRPS, the adoption of path smoothing is beneficial for a smaller path cost $\mathcal{C}$ but results in a much longer path length, even with the computation time for post-processing three times that for A*.

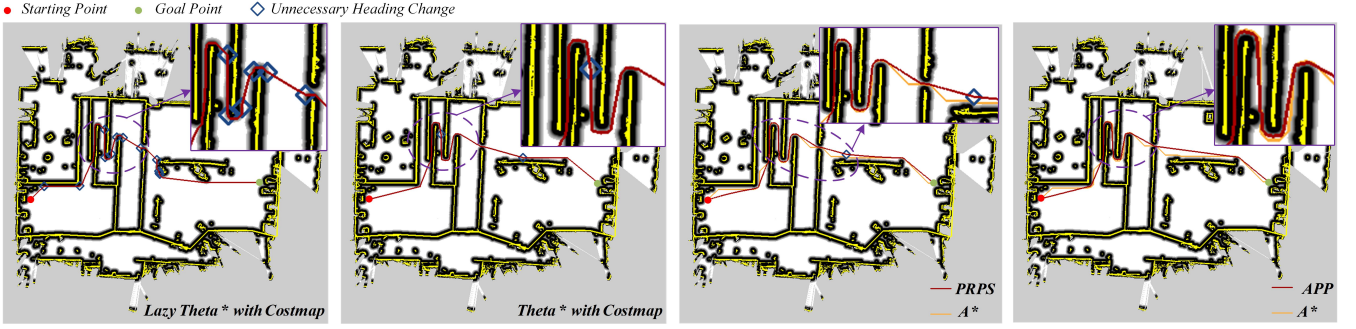

Fig. 9. Comparative experimental results in experiment III. Parts of the paths are enlarged to demonstrate the details better.

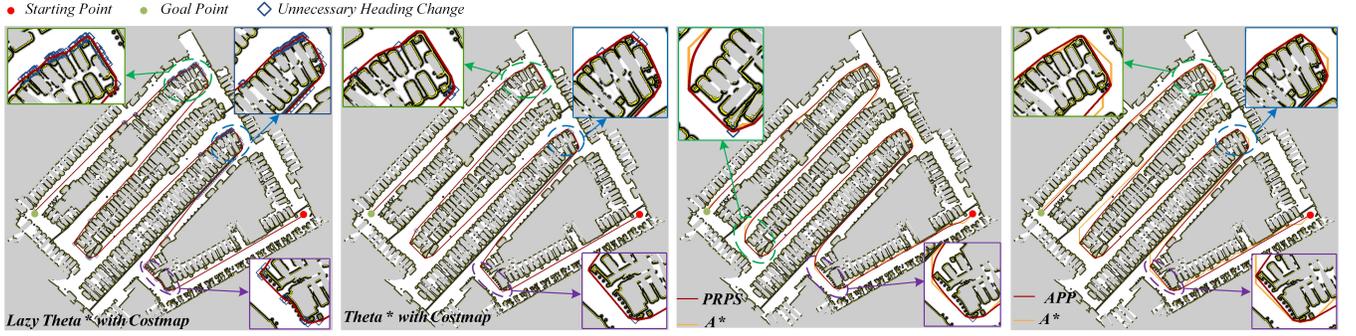

Fig. 10. Comparative experimental results in experiment IV. Parts of the paths are enlarged to demonstrate the details better.

TABLE I. PERFORMANCE INDEXES IN EXPERIMENT I-IV

|   |   | $t$ | $l$ | $\mathcal{C}$ | $\mathcal{N}_\theta$ | $\mathcal{S}$ |
|---|---|---|---|---|---|---|
| Exp. I | Lazy Theta* | 0.270 | 1607.070 | 44.359 | 25 | 1.598 |
|   | Theta* | 0.262 | 1599.070 | 19.913 | 3 | 1.544 |
|   | PRPS | 0.859 | 1611.73 | **12.974** | 3 | 0.219 |
|   | **APP** | **0.259** | **1566.730** | 17.304 | **0** | **0.0796** |
| Exp. II | Lazy Theta* | 0.335 | 1519.850 | 64.168 | 63 | 3.895 |
|   | Theta* | 0.285 | 1513.270 | 24.024 | 3 | 0.322 |
|   | PRPS | 0.94 | 1505.500 | **14.757** | 2 | 0.279 |
|   | **APP** | **0.247** | **1480.230** | 24.263 | **0** | **0.218** |
| Exp. III | Lazy Theta* | 0.445 | 852.570 | 90.539 | 12 | 2.368 |
|   | Theta* | 0.338 | 843.037 | 39.734 | 2 | 1.471 |
|   | PRPS | 0.563 | 849.166 | **32.334** | 1 | **0.983** |
|   | **APP** | **0.267** | **809.965** | 35.431 | **0** | 1.342 |
| Exp. IV | Lazy Theta* | 17.383 | 9655.870 | 29.095 | 120 | 1.385 |
|   | Theta* | 12.611 | 9636.980 | 10.410 | 9 | 0.126 |
|   | PRPS | 8.096 | 9687.25 | **8.793** | 2 | 0.119 |
|   | **APP** | **5.593** | **9622.410** | 11.660 | **0** | **0.0978** |

**Experiment II**: The experimental environment is the lobby of CVTE with the map grid size of 1440 × 1060. Comparative experimental results and performance indexes are shown in Fig. 8 and Table I, respectively. The resulting path of APP does not have any unnecessary heading changes. In contrast, both Lazy Theta* and Theta* have unnecessary heading changes in free space, which is inconsistent with people's common sense that the Point-To-Point path should be straight when there are no obstacles nearby. Compare Fig. 4 with 8 and we can see that: after the first iteration of vertices reduction in APP, the resulting path still has some unnecessary heading changes. However, an improved post-processing performance for APP is achieved after iterative bidirectional vertices reduction and path perturbation. Similar experimental results for PRPS are obtained in Experimental I and II. When the iteration upper bound for path smoothing is selected as 30,000, the path for PRPS is similar with that of APP in vision, with the planning time $t$ 6.245s, which restricts the application of PRPS on computational-ability-constrained computation units.

**Experiment III**: The experimental environment has several complex narrow passages and 180° U-turns, with the costs of some girds higher than $\varepsilon_{cost}$. The map grid size is 1440 × 1060. Comparative experimental results and performance indexes are shown in Fig. 9 and Table I, respectively. For APP, the vertices with costs higher than cost threshold $\varepsilon_{cost}$ remain unchanged during vertices reduction and path perturbation, which guarantees the safety of the path. Path relaxation for PRPS moves path points down the cost surface to increase clear-up, resulting in smaller $\mathcal{C}$ and $\mathcal{S}$, especially in narrow passages.

**Experiment IV**: The experimental environment is an over-sized garage with the map grid size of 3040 × 2820. Some passages are blocked manually to make the map a large and complex corridor-like environment. Comparative experimental results and performance indexes are shown in Fig. 10 and Table I, respectively. Compared with Lazy Theta*, Theta*, and PRPS, the planning time of APP has been decreased by 55.6%, 67.8% and 30.92%, respectively, which verifies the computational efficiency of APP in complex environments. Besides, we can notice that the computation time of Lazy Theta* is longer than that of Theta*, which is maybe resulted from the inflated costmap with obstacles exceeding 30%. Path cost $\mathcal{C}$ for APP is larger than Theta* and PRPS and this is because: APP has a much better performance in path-shortening, resulting in the costs of some path vertex the line-of-sight-check cost threshold $\varepsilon_{cost}$, e.g., path sequences in the enlarged windows in Fig.10.

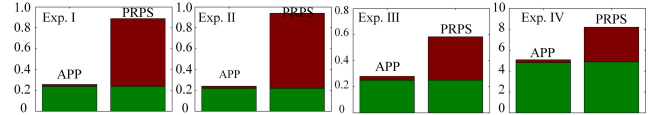

Fig. 11. Contribution analysis of computation time for APP and PRPS in experiment I-IV. The green and red bars represent the computation time for A* and post-processing, respectively. For APP, the post-processing process needs only a little extra computation time.

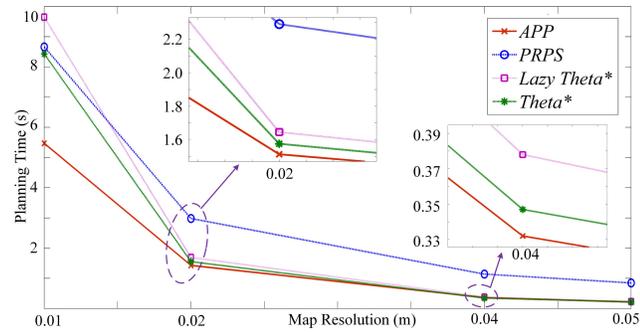

Fig. 12. Computation time comparison with the map resolution 0.05, 0.04, 0.02 and 0.01 for the environment in Experiment I.

For APP and PRPS, the performance index $t$ is the total planning time, including A* and post-processing. Fig. 11 shows the comparative contributions of computation time for A* and post-processing in Experiments I-IV: APP needs a little extra computation time to improve post-processing performance, especially in path length and unnecessary heading changes. Actually, the basic flow for path perturbation algorithm in APP is similar to that for the path smoothing algorithm in [15]: Vertex $p_i$ is perturbed based on its predecessor $p_{i-1}$ and successor $p_{i+1}$ with the computation complexity $O(n^2)$. The main difference between APP and PRPS is that, for APP, path perturbation is adopted only to fine-tune the path with a small iteration number of 20, and the guaranteed path-shortening performance is achieved by the efficient bidirectional vertex reduction algorithm.

Experiments I-IV are conducted with the map resolution 0.05. To further compare the computation efficiencies of the four methods in more dense maps, comparative experiments are designed. The laser and odometry data during the mapping process for the environment in Experiment I have been recorded. And we replay the data to produce maps that are with the resolution of 0.01, 0.02 and 0.04. Comparative exp-erimental results in Fig. 12 show that APP outperforms other methods both in low and high resolution maps.

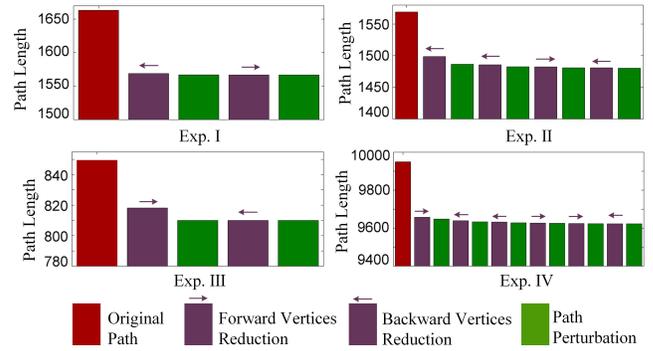

Fig. 13. Evolution of path lengths for APP in experiment I-IV.

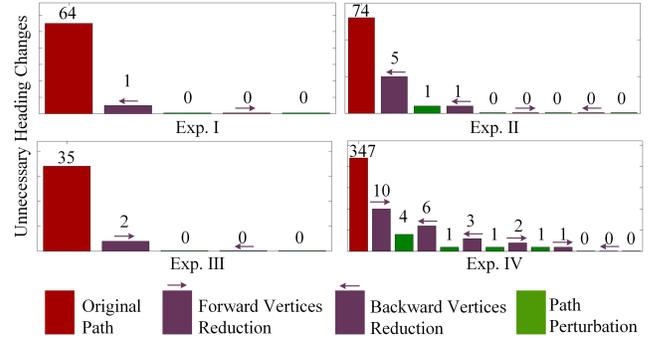

Fig. 14. Evolution of unnecessary heading changes for APP in experiment I-IV. The heights of the bars for original paths are reduced for a better comparative display effect, and the numbers on the top of each bar indicate the actual unnecessary heading changes after a specific process, i.e., forward vertices reduction, backward vertices reduction and path perturbation.

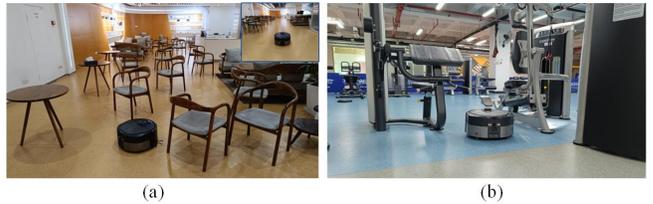

Fig. 15. Field navigation experiments in complex commercial environments: (a) a cafe with randomly placed chairs blocking the way; (b) a gymnasium with suspended and low obstacles.

To further analyze the contributions of the bidirectional vertices reduction and path perturbation in post-processing, the evolution process for path lengths and unnecessary heading changes in Experiment I-IV are shown in Fig. 13 and Fig. 14, respectively. Vertices reduction for the first iteration contributes most to path-shortening and unnecessary-heading-changes-eliminating. Besides, the vertices reduction algorithm removes unnecessary vertices for the path-shortening process. Then specific vertices will be locally perturbed to improve the straightness of the path, which guarantees the asymptotic convergence of path length in Fig. 13. It is interesting to find from Fig. 14 that: In experiment IV, $\mathcal{N}_\theta$ increases after path perturbation for the first three iterations (with $\mathcal{N}_\theta$ changing from 4 to 6, 1 to 3 and 1 to 2). It is because for a vertex $p_i$ with unnecessary heading change, path perturbation at $p_i$ eliminates its unnecessary heading change but may bring in that at $p_{i-1}$ and $p_{i+1}$. This problem can be solved by adopting a larger $n_{sm}$, or a large but dynamically adjusted $n_{sm}$, considering the computational efficiency for a long path.

*B. Field experiments with commercial cleaning robots*

To verify the applicability of APP to dynamic and confined environments, field autonomous navigation experiments are carried out with a low-cost and computational-ability-constrained commercial cleaning robot, produced by CVTE with RK3399 the computation unit. The global path planner (A* + APP) runs offline to generate a global reference path. When the path is predicted to collide with obstacles, the local path planner (A* + APP) will replan a safer path back to the global one. Paths generated by global and local path planners will be curve-fitted by a non-uniform B-Spline to provide reference velocities for a kinematic-based nonlinear motion controller.

Field Experiment I (Fig. 15(a)) is conduct in a cafe. The space between the start and the goal is free (seen upper right corner of Fig. 15(a)) and the resulting global path is a straight line. During the navigation experiment, the authors randomly placed some chairs with thin legs to block the way. The robot can still pass through the "chair forest" and navigate to the goal safely. Field Experiment II (Fig. 15(b)) is carried out in a gymnasium. The APP-based global path planner generates a path traversing through the gymnasium. The suspended and lower parts of the fitness equipment are detected by a RGBD camera and the APP-based local path planner guarantees the safeness during the autonomous navigation. Videos for Field Experiment I-II are submitted as supplementary materials.

## V. CONCLUSION

This article puts forward a general and systematic post-processing algorithm for A* and other graph-search-based planners. The thorough-shortcut-based bidirectional vertices reduction achieves guaranteed path-shortening performance, and the introduced path perturbation strategy achieves improved performances in path smoothness. Comparative experimental results show that APP achieves superior performances in planning time (especially in large and complex environments) and path length. The numbers of unnecessary heading changes for APP are zero in four comparative experiments, which means that the resulting paths are more understandable and acceptable to humans.

Our researches focus on solving the critical motion planning problems that prevent commercializing robots in unstructured and semi-structured environments. Research on fast global path planning for large maps and optimal-time trajectory planning will be carried out in the future.

## REFERENCES

[1] R. Bormann, F. Jordan, J. Hampp and M. Hägele, "Indoor Coverage Path Planning: Survey, Implementation, Analysis," in 2018 *IEEE International Conference on Robotics and Automation (ICRA)*. IEEE, 2018, pp. 1718–1725.

[2] Y. Li and H. Cheng, "Unidirectional-Road-Network-Based Global Path Planning for Cleaning Robots in Semi-Structured Environments," in 2023 *IEEE International Conference on Robotics and Automation (ICRA)*. IEEE, 2023, pp. 1572–1578.

[3] M. M. Rayguru, M. R. Elara, A. A. Hayat and B. Ramalingam, and S. Roy, "Modeling and control of panthera self-reconfigurable pavement sweeping robot under actuator constraints," in 2021 *IEEE/RSJ International Conference on Intelligent Robots and Systems (IROS)*. IEEE, 2021, pp. 2742–2748.

[4] D. González, J. Pérez, V. Milanés and F. Nashashibi, "A review of motion planning techniques for automated vehicles," *IEEE Transactions on Intelligent Transportation Systems*, vol. 17, no. 4, pp. 1135–1145, 2016.

[5] S. Kim and M. Likhachev, "Path Planning for a Tethered Robot using Multi-Heuristic A* with Topology-based Heuristics," in 2015 *IEEE/RSJ International Conference on Intelligent Robots and Systems (IROS)*. IEEE, 2015, pp. 4656–4663.

[6] J. Ha and S. Kim, "Fast Replanning Multi-Heuristic A," in 2021 *IEEE International Conference on Robotics and Automation (ICRA)*. IEEE, 2021, pp. 7430–7435.

[7] P. Hart, N. Nilsson and B. Raphael. "A formal basis for the heuristic determination of minimum cost paths," *IEEE Transactions on Systems Science and Cybernetics*, vol. 4, no. 2, pp. 100–107, 1968.

[8] K. Daniel, A. Nash, S. Koenig and A. Felner, "Theta*: Any-angle path planning on grids," *Journal of Artificial Intelligence Research*, vol. 39, pp. 533–579, 2010.

[9] A. Nash, S. Koenig and C. Tovey, "Lazy Theta*: Any-angle path planning and path length analysis in 3D," In *Proceedings of the AAAI Conference on Artificial Intelligence (AAAI)*, pp. 147–154, 2010.

[10] S. Choi and W. Yu, " Any-angle Path Planning on Non-uniform Costmaps," in 2011 *IEEE International Conference on Robotics and Automation (ICRA)*. IEEE, 2011, pp. 5615–5621.

[11] H. Kim, T. Lee, C. Hyun, et al., "Any-angle path planning with limit-cycle circle set for marine surface vehicle," in 2012 *IEEE International Conference on Robotics and Automation (ICRA)*, IEEE, 2012, pp. 2275–2280.

[12] M. Kanehara, S. Kagami, J. J. Kuffner, S. Thompson, and H. Mizoguhi, "Path shortening and smoothing of grid-based path planning with consideration of obstacles," in *Proceedings of IEEE International Conference on Systems, Man, and Cybernetics (SMC)*, IEEE, 2007, pp. 991–996.

[13] A. Botea, M. Muller and J. Schaeffer, "Near optimal hierarchical path-finding," *Journal of Game Development*, vol. 1, no. 1, pp. 1–22, 2006.

[14] D. Ferguson and A. Stentz, "Using interpolation to improve path planning: The Field D* algorithm," *Journal of Field Robotics*, vol. 23, no. 2, pp. 79–101, 2006.

[15] A. Richardson and E. Olson, "Iterative path optimization for practical robot planning," In 2011 *IEEE/RSJ International Conference on Intelligent Robots and Systems(IROS)*. IEEE, 2011, PP.3881–3886.

[16] R. Geraerts, and M. H, Overmars, "Creating high-quality paths for motion planning," *The international journal of robotics research*, vol. 26, no. 8, 845–863, 2007.

[17] I. A. Sucan, M. Moll, and L. E. Kavraki, "The Open Motion Planning Library," *IEEE Robotics & Automation Magazine*, vol. 19, no. 4, pp. 72–82, 2012.

[18] E. Heiden, L. Palmieri, S. Koenig, K. O. Arras and G. S. Sukhatme, "Gradient-informed path smoothing for wheeled mobile robots," in 2018 *IEEE International Conference on Robotics and Automation (ICRA)*, IEEE, 2018, pp. 1710–1717.

[19] T. W. Kang, J. G. Kang and J. W. Jung, "A Bidirectional Interpolation Method for Post-Processing in Sampling-Based Robot Path Planning," *Sensors*, vol. 21, no. 21, pp. 7425, 2021.

[20] M. Quigley, B. Gerkey, K. Conley, J. Faust, T. Foote, J. Leibs, E. Berger, R. Wheeler, and A. Ng, "ROS: an open-source Robot Operating System," in ICRA Workshop on Open Source Software, Kobe, Japan, 2009.

[21] E. Marder-Eppstein, E. Berger, T. Foote, B. Gerkey and K. Konolige, "The office marathon: Robust navigation in an indoor office environment," In 2010 IEEE international conference on robotics and automation (ICRA). IEEE, 2010, pp. 300–307.

[22] D. V. Lu, D. Hershberger and W. D. Smart, "Layered costmaps for context-sensitive navigation," In 2014 *IEEE/RSJ International Conference on Intelligent Robots and Systems(IROS)*. IEEE, 2014, pp. 709–715.